\documentclass[11pt]{article}
\usepackage{preprint}          
\makeatletter
\providecommand\@LN[2]{}
\providecommand\@LN@col[1]{}
\makeatother

\usepackage{inconsolata}
\usepackage{flafter}
\usepackage{tabularx}
\usepackage{tikz}
\usetikzlibrary{arrows.meta}
\usepackage{fontawesome5}
\newcommand{\yuhan}[1]{#1}

\definecolor{ablationhl}{HTML}{E2F0E4}
\definecolor{retentionhl}{HTML}{F3F0E8}
\definecolor{casepurple}{HTML}{69649A}
\definecolor{caseblue}{HTML}{255C99}
\definecolor{caseorange}{HTML}{AA5F22}
\definecolor{casegreen}{HTML}{23704C}
\usetikzlibrary{patterns.meta}
\newcommand{\casefact}[1]{\noindent\textbullet\ #1\par}
\newcommand{\casehyp}[3]{
\noindent\begin{tabularx}{\linewidth}{@{}l@{\hspace{3pt}}X@{\hspace{4pt}}l@{}}
\textcolor{casegreen}{$\delta_{#1}$} & #2 &
\begin{tikzpicture}[baseline=-0.5ex]
\fill[casegreen!12] (0,0) rectangle (1.0,0.16);
\fill[casegreen] (0,0) rectangle ({1.0*#3},0.16);
\end{tikzpicture}
\end{tabularx}\par
}
\newtcolorbox{casestage}[2]{enhanced,colback=white,colframe=#1!45,
  colbacktitle=#1!8,coltitle=#1,boxrule=0.5pt,arc=1.2mm,
  left=2.5mm,right=2.5mm,top=2mm,bottom=2mm,
  before skip=0pt,after skip=0pt,fonttitle=\bfseries,title={#2}}
\newtcolorbox{caselatent}[2]{enhanced,colback=black!14!white,
  colframe=#1!28!white,boxrule=0.35pt,arc=0.8mm,boxsep=0pt,
  left=1.3mm,right=1.3mm,top=1mm,bottom=1mm,
  before skip=0pt,after skip=0pt,valign=top,
  equal height group=case-latent-#2,
  underlay={\begin{tcbclipinterior}
    \path[pattern={Hatch[angle=45,distance=2mm,line width=0.18pt]},
      pattern color=black!28!white]
      (interior.south west) rectangle (interior.north east);
    \end{tcbclipinterior}},
  shadow={1.2mm}{-1.2mm}{0mm}{fill=black!35!white,draw=none}}
\newcommand{\caseinteraction}[3]{
\noindent\textcolor{caseblue}{\textbf{Step #1}}\quad\textbf{Action:} \texttt{#2}\par
\noindent\textbf{Observation:} \textit{#3}\par\smallskip
}
\newtcolorbox{casepanel}[1]{enhanced,colback=white,colframe=casepurple,
  colbacktitle=casepurple,coltitle=white,boxrule=0.8pt,arc=2mm,
  left=3mm,right=3mm,top=2.5mm,bottom=2.5mm,
  before skip=0pt,after skip=0pt,fonttitle=\bfseries\large,title={#1}}

\title{LAMB: Latent Multi-Hypothesis Beliefs\\for Long-Horizon Large Language Agents}
\shorttitle{LAMB: Latent Multi-Hypothesis Beliefs for Long-Horizon Language Agents}
\newcommand{\method}{LAMB}

\author{Peien Ding$^{1}$\thanks{Equal contribution.} \and Hui Li$^{1}$\footnotemark[1] \and Yuhan Liu$^{2}$\thanks{Corresponding authors.} \and Jinsong Su$^{1}$\footnotemark[2]}
\affiliations{$^{1}$School of Informatics, Xiamen University\quad $^{2}$MBZUAI}
\correspondence{\href{mailto:pnding@stu.xmu.edu.cn}{pnding@stu.xmu.edu.cn}\quad \href{mailto:yuhan.liu@mbzuai.ac.ae}{yuhan.liu@mbzuai.ac.ae}\quad \href{mailto:jssu@xmu.edu.cn}{jssu@xmu.edu.cn}}
\preprintnote{Preprint.}

\begin{document}
\maketitle

\begin{abstract}
\yuhan{Long-horizon language agents require compact memories that retain
interaction evidence while preserving competing hypotheses about
unobserved states.
Although effective, many belief-state methods encode memory in
discrete natural language, where repeated compression can omit
evidence or prematurely discard plausible alternatives. We introduce \method{}, a recurrent framework for latent multi-hypothesis
beliefs. A Common--Delta factorization combines a shared Common state for
interaction evidence with weighted Delta components for competing
hypotheses.
Relative evidence from action-outcome predictions guides recurrent
hypothesis updates without generating textual memory.
A context-conditioned readout exposes hypotheses separately to
the policy through a fixed-token belief prefix in the default
configuration.
The belief modules and policy are jointly trained using action
imitation and auxiliary state, outcome, and observation-discrimination
objectives.
Across ALFWorld, BabyAI, ScienceWorld, and Crafter, \method{} matches
or outperforms PABU, improving ALFWorld SR@30 and ScienceWorld
success by 6.0 and 22.0 percentage points, respectively, and Crafter
achievement progress by 26.37 points.
On ALFWorld, it also reduces the redundant-action rate by
17.70 percentage points.
These results support latent multi-hypothesis beliefs as compact
memory for effective long-horizon decision making.}
\end{abstract}

\begin{figure}[!t]
\centering
\includegraphics[width=0.78\linewidth]{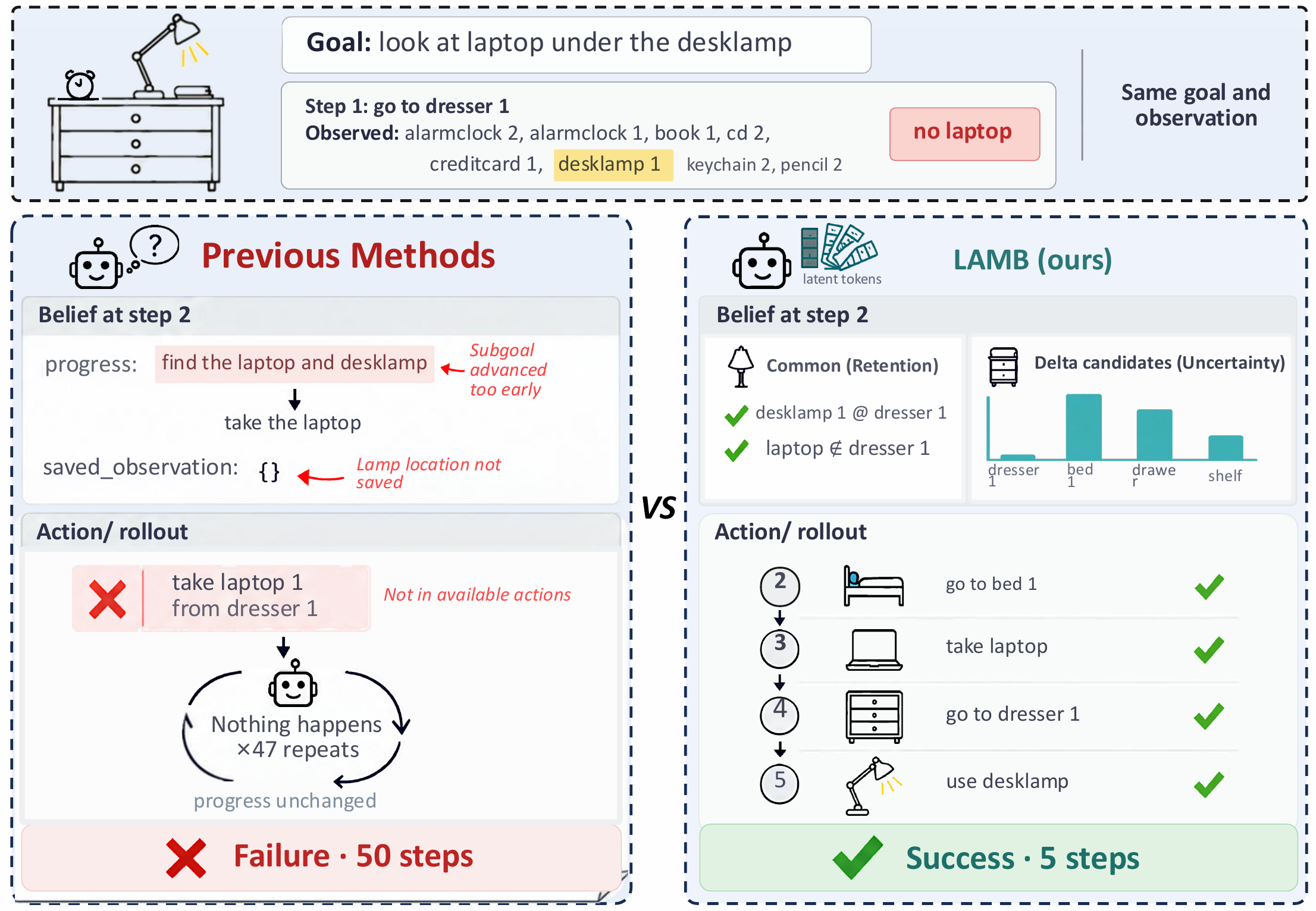}
\caption{Existing text-based beliefs (left) prematurely drop or misjudge environmental observations, trapping the agent in invalid action loops. 
\method{} (right) factorizes beliefs into confirmed state facts and candidate spatial distributions, maintaining calibrated uncertainty to resolve the task efficiently in 5 steps. }
\label{fig:intro-belief-comparison}
\end{figure}

\section{Introduction}\label{introduction}

\yuhan{Large language model (LLM) agents solve multi-step tasks through
sequences of actions and observations~\citep{yao2023react},
which has wide applications in many fields~\citep{liu2024skepticism,liu2025stepwise,liu2025truth}.
Under partial observability, effective decisions require retaining
earlier evidence while reasoning about aspects of the environment
that remain unobserved.
Conditioning on the full interaction history provides access to
past information, but its growing length increases the context
processed at each decision.
Long-horizon agents therefore need memory that preserves
decision-relevant information without repeatedly processing
the complete trajectory.}

\yuhan{Context-management methods reduce this burden by pruning,
masking, or compressing past
interactions~\citep{lindenbauer2025complexity,xiao2025agentdiet}.
Belief-state methods instead recursively update a compact state
for subsequent decisions.
Although simple and effective, many LLM belief-state methods
still represent beliefs in discrete natural
language~\citep{pabu,abbel,agentbrace}.
Even when augmented with confidence labels, these states must
express retained evidence and alternatives within a limited
text budget.
Repeated selection and rewriting can omit earlier findings
or prematurely discard plausible interpretations.

For example, Figure~\ref{fig:intro-belief-comparison} shows an agent ruling
out the dresser while considering the bed, drawer, and shelf
as possible laptop locations.
Its memory should preserve the inspection result shared by
both alternatives and revise the candidates as new evidence arrives.
Updating textual beliefs through generation also incurs decoding
overhead beyond action selection.
These considerations motivate our central question:
\emph{how can a fixed-size recurrent memory retain shared evidence
while preserving and revising competing hypotheses without
repeatedly verbalizing its state?}}

\yuhan{To address this question, we propose \method{}, a latent
belief framework that tracks episode information directly
in continuous latent space.
Through a Common--Delta factorization, a shared Common component
is trained to retain interaction evidence, while weighted Delta
components encode hypothesis-specific information.
Rather than pooling the particles into a single representation,
a context-conditioned readout exposes each Delta to the policy
separately, alongside the Common readout.
The default configuration produces a fixed-token belief prefix,
keeping the historical contribution to policy input constant
throughout the episode.
To make hypothesis revision responsive to new evidence, \method{}
compares each particle's action-outcome prediction with the
feedback received from the environment.
The resulting relative evidence guides recurrent Delta updates,
after which mixture weights are recomputed from the updated
Common and Delta states.
We jointly train the belief modules and policy on offline
interaction trajectories using action imitation, joint
state-semantic prediction, outcome-semantic prediction, and
observation discrimination for the Common component.
Simulator-derived state labels are used only during training;
online belief updates rely on observed actions and feedback.}

\yuhan{We evaluate \method{} on ALFWorld, BabyAI, ScienceWorld, and Crafter.
Results show that \method{} matches or outperforms the strong baselines across
these benchmarks while reducing token consumption on ALFWorld.
Ablations and diagnostic analyses further assess the proposed
design and characterize evidence retention, particle readouts,
and agent behavior.}

Our main contributions are as follows:

\begin{itemize}[leftmargin=*]
\item
  We propose \method{}, a fixed-capacity latent belief framework
    that supports long-horizon decisions with a constant-size
    policy prefix, without generating textual memory updates.
\item
  We develop a Common--Delta representation with action-outcome
    prediction, evidence-guided recurrent updates, and an unpooled
    readout of weighted hypotheses.
\item
  \yuhan{We demonstrate task-performance gain across four benchmarks and inference cost on ALFWorld}, complemented by ablations, redundant-action analysis,  \yuhan{and diagnostics of evidence
    retention and particle readouts.}
\end{itemize}

\section{Related Work}\label{related-work}

\subsection{Stream-based Agent Memory}

\yuhan{Many agent-memory systems organize past interactions as records that are retrieved, revised, or compressed for subsequent use.} Generative Agents~\citep{park2023generative}
and MemGPT~\citep{packer2023memgpt} keep the full record in external storage
and page relevant entries into context, \yuhan{A-MEM~\citep{xu2025amem}
links and updates memory entries.}  \yuhan{Compression methods, including} LLMLingua-2~\citep{pan2024llmlingua2}, LongLLMLingua~\citep{jiang2024longllmlingua},
and ACON~\citep{kang2025acon} \yuhan{reduce the context passed to the policy.} Experience-reuse methods, including Reflexion~\citep{shinn2023reflexion},
ExpeL~\citep{zhao2024expel}, AWM~\citep{wang2024awm}, and ACE~\citep{zhang2025ace},
distill cross-episode knowledge into insights, workflows, or playbooks.

\yuhan{These approaches differ in their retention and access mechanisms; the accessibility of retained evidence can also depend on its placement within the context~\citep{liu2024lost}.}

\yuhan{Our \method{} focuses on maintaining a fixed-size state of the current episode through recursive updates, which offers a different interface for carrying interaction evidence into each decision.}
\subsection{\yuhan{Textual Belief State}}

\yuhan{Belief-state approaches maintain a compact representation for decision making under partial observability~\citep{kaelbling1998pomdp}.}

\yuhan{MEM1~\citep{zhou2025mem1} and ABBEL~\citep{abbel} learn textual states that consolidate information across interactions, while PABU~\citep{pabu} tracks task progress together with selectively retained actions and observations.}
Agent-BRACE~\citep{agentbrace}
and BeliefMem~\citep{beliefmem} further encode uncertainty through
confidence labels or candidate records.
Textual beliefs, however, represent uncertainty by \emph{enumeration}:
evidence shared across hypotheses is restated for each candidate, so a
bounded state must trade evidence retention against hypothesis coverage.
In practice this trade-off surfaces either as forgotten findings or as
premature commitment to a single hypothesis~\citep{krishnamurthy2024explore}.
\method{} keeps hypotheses in latent form on top of a shared common state,
so each additional hypothesis costs one prefix token rather than another
restatement of the evidence.

\subsection{\yuhan{Latent Memory and Filtering}}

\yuhan{Latent-memory methods encode context as learned continuous representations, avoiding explicit text generation for the memory itself.}
Recurrent Memory
Transformer~\citep{bulatov2022rmt}, AutoCompressors~\citep{chevalier2023autocompressors},
Gist tokens~\citep{mu2023gist}, Laser~\citep{wang2026forest} and ICAE~\citep{ge2024icae} condense context
into a small set of learned vectors. \yuhan{In agent settings,} Mem-W~\citep{memw} and
MemGen~\citep{zhang2025memgen} bring such representations to agents as
within-episode and generative memory. 

However, particle-based belief models, such as Particle Filter Networks~\citep{karkus2018pfnet},
DVRL~\citep{igl2018dvrl}, and PF-RNN~\citep{ma2020pfrnn}, make hypotheses
explicit, but each particle \emph{replicates} the full state, reintroducing
the duplication of shared evidence.
In contrast, \method{} factorizes a recurrent latent belief into a common
component that stores shared evidence once and hypothesis-specific deltas
that encode only how alternatives differ.

\section{Method}\label{method}

\begin{figure}[t]
\centering
\includegraphics[width=\textwidth]{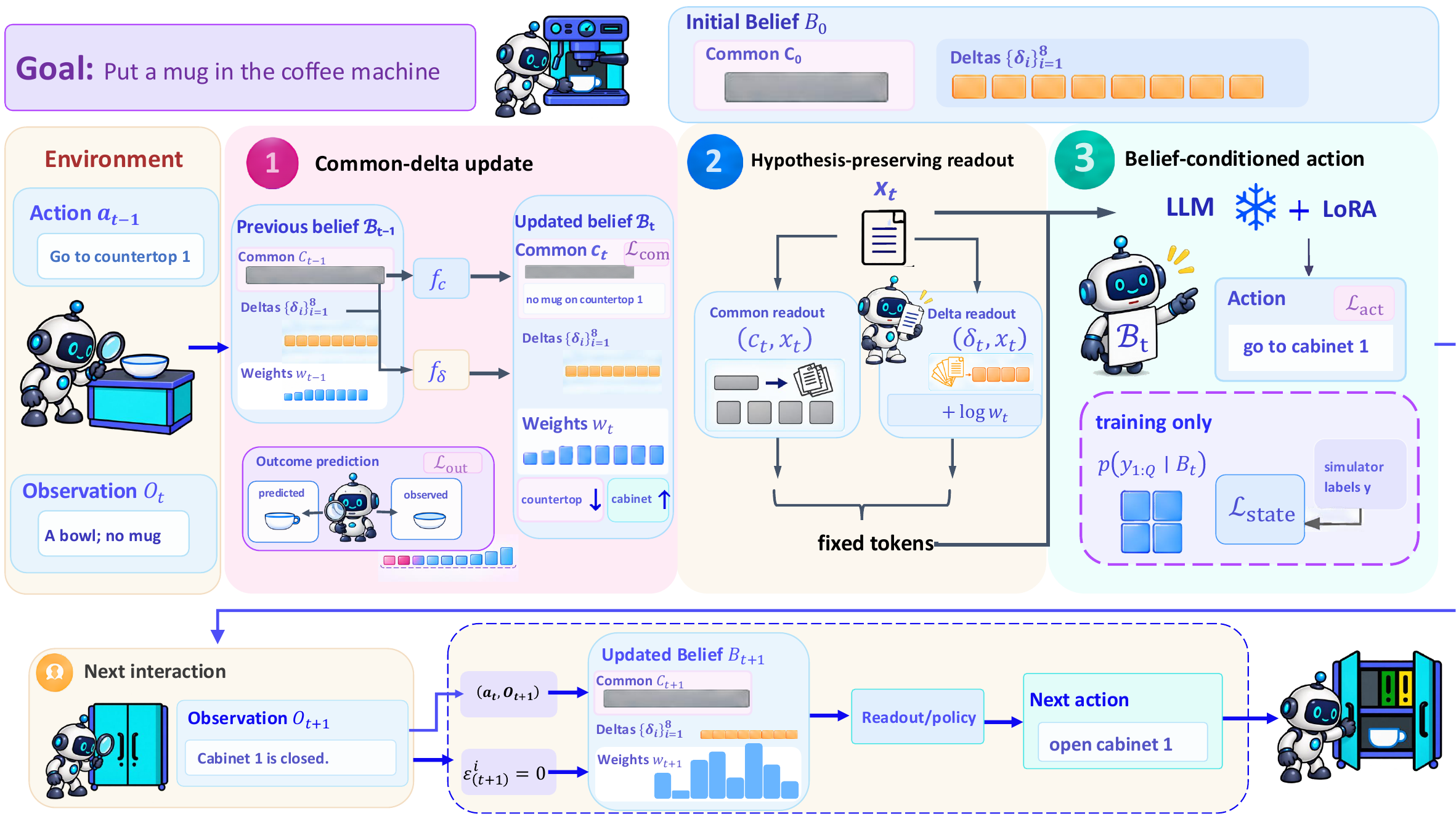}
\caption{Overview of \method{}. The recurrent updater incorporates the latest action and observation into a shared Common state and weighted Delta hypotheses. A context-conditioned readout produces a fixed 12-token prefix for action generation. Simulator facts supervise joint prediction during training only; subsequent interaction feedback enters the next belief update.}
\label{fig:overview}
\end{figure}

\subsection{\yuhan{Factorizing the Latent Belief}}
\label{sec:formulation}

\yuhan{We consider goal $g$-directed interaction under partial observability.}
At step $t$, the agent receives an observation $o_t$ and selects an action
$a_t$ from the admissible set $\mathcal{A}_t$. A history-conditioned policy
acts on:
\begin{equation}
    h_t=(g,o_1,a_1,\dots,a_{t-1},o_t),a_t\sim\pi_\theta(\cdot\mid h_t), 
\end{equation}
whose input length grows linearly with $t$.
\yuhan{Our method \method{} replaces the history before the current step with a
fixed-size belief $B_t$, updated from the latest action and observation:
\begin{equation}
    \begin{aligned}
        B_0 = \operatorname{Init}(g),
        B_t = U\bigl(B_{t-1},a_{t-1},o_t\bigr).
    \end{aligned}
    \label{eq:pb-recurrent-belief}
\end{equation}
The policy conditions on the updated belief and the current context
$x_t=(g,o_t,\mathcal{A}_t,a_{t-1})$:
\begin{equation}
    a_t \sim \pi_\theta\bigl(\cdot \mid R(B_t,x_t),x_t\bigr),
    \label{eq:pb-belief-policy}
\end{equation}
where $R$ maps the belief to a fixed number of continuous prefix tokens.
The historical contribution to policy input therefore remains constant,
while the current observation and candidate actions may vary in length.}

\yuhan{Partial observations can leave several world configurations plausible.
To share information across these alternatives while allowing
particle-specific representations, we factorize the belief into a
common state $c_t$, $K$ delta states $\{\delta_t^i\}_{i=1}^{K}$, and
normalized mixture weights $w_t \in \Delta^{K-1}$:
\begin{equation}
    B_t = \bigl(c_t,\{\delta_t^i\}_{i=1}^{K},w_t\bigr),\quad
    c_t,\delta_t^i \in \mathbb{R}^{d}.
    \label{eq:pb-common-delta}
\end{equation}
The common state is intended to retain information useful across
hypotheses, while the deltas provide separate representations that can
capture unresolved alternatives. Each pair $(c_t,\delta_t^i)$
constitutes a learned latent particle. This factorization encourages
information sharing and component-specific variation without imposing
a uniquely identifiable semantic decomposition.}

\yuhan{During training, the belief predicts a set of simulator-derived fact
values $y_{1:Q}$ through a mixture of conditionally independent
distributions:
\begin{equation}
    p\bigl(y_{1:Q}\mid B_t\bigr)
    = \sum_{i=1}^{K} w_t^i
      \prod_{q=1}^{Q}
      p_\phi\bigl(y_q\mid c_t,\delta_t^i\bigr).
    \label{eq:mixture}
\end{equation}
Each component factorizes the fact predictions, while the shared
mixture index couples them in the joint distribution. This construction
can capture dependencies such as mutually exclusive object locations.
The resulting consistency depends on the learned component predictions;
the mixture form alone does not exclude invalid configurations
(Appendix~\ref{app:joint-prediction}).}

\yuhan{Figure~\ref{fig:overview} summarizes the interaction loop. The updater evaluates the
observed action outcome under the previous particles, updates the
common and delta states, and recomputes the mixture weights
(Section~\ref{sec:update}). A context-conditioned readout then
converts the updated belief into a fixed-length prefix for action
generation (Section~\ref{sec:readout}). The policy and belief modules
are trained jointly (Section~\ref{sec:training}).
}

\subsection{Common-Delta Belief Update}
\label{sec:update}

Updating beliefs under partial observability requires resolving a fundamental \yuhan{problem: \textit{committing to verified consensus across the episode while preserving distinct, competing hypotheses where ambiguity remains.}} To balance shared facts against hypothesis diversity, our update decouples belief evolution into an objective common state $c_t$ and specialized particle deltas $\delta_t^i$. 
At each step, the belief update incorporates the latest transition $(a_{t-1}, o_t)$ to evaluate competing hypotheses and evolve the state representations. We first encode the transition into $u_t = \operatorname{Embed}(a_{t-1}, o_t)$. Before modifying the state vectors, \method{} evaluates how accurately each hypothesis anticipated the transition outcome $r_t$:
\begin{equation}
\begin{aligned}
\epsilon_t^i &= \ell_t^i - \frac{1}{K}\sum_{j=1}^K \ell_t^j, \\
\ell_t^i &= \log p_\psi\big(r_t \mid c_{t-1}, \delta_{t-1}^i, a_{t-1}\big).
\end{aligned}
\label{eq:evidence}
\end{equation}
Here, $\epsilon_t^i$ serves as a relative evidence signal: hypotheses whose predictions better match the observed transition receive positive feedback, while conflicting ones are penalized.

Next, both components evolve via gated residual updates:
\begin{align}
c_t &= \mathrm{Gate}_c\big(c_{t-1}, [c_{t-1}; u_t]\big), \\
\delta_t^i &= \mathrm{Gate}_\delta\big(\delta_{t-1}^i, [c_{t-1}; \delta_{t-1}^i; u_t; \epsilon_t^i]\big),
\end{align}
where $\mathrm{Gate}(s, z) = (1 - \sigma(W_z z)) \odot s + \sigma(W_z z) \odot \operatorname{MLP}(z)$. 
Crucially, the information flow is asymmetric: the common state $c_t$ aggregates objective transition history without consulting individual hypotheses, while each delta $\delta_t^i$ conditions on both $c_{t-1}$ and its specific evidence $\epsilon_t^i$. \yuhan{We parameterize $\mathrm{Gate}_\delta$ with a per-hypothesis LoRA~\citep{hu2022lora} to encourage specialization across particles.
Finally, a shared scoring head $g$ re-estimates the belief weights directly from the updated states:
\begin{equation}
w_t^i = \frac{\exp\big(g(c_t, \delta_t^i)\big)}{\sum_{j=1}^K \exp\big(g(c_t, \delta_t^j)\big)}.
\end{equation}}

\subsection{Unpooled Policy Readout}
\label{sec:readout}

Directly pooling particle beliefs into a single vector collapses multimodal uncertainty, blending mutually exclusive hypotheses into ambiguous mean states. To preserve discrete alternatives while anchoring the decision in established facts, we design an asymmetric readout that translates the belief into a structured prefix sequence.

Specifically, the common state $c_t$ is integrated with the current input representation $\operatorname{Embed}(x_t)$ via cross-attention into $M$ context tokens, capturing verified environmental constraints. In parallel, each particle $\delta_t^i$ is projected into an individual hypothesis token $e_t^i$, modulated by its log-weight: $\tilde{e}_t^i = e_t^i + v \log w_t^i$, where $v$ is a learned direction vector. 
Maintaining $K$ separate hypothesis tokens enables the policy's self-attention mechanism to explicitly contrast candidate states, isolating dominant hypotheses from marginal ones without information loss. Finally, the resulting $M+K$ belief tokens are prepended as a belief-conditioned continuous prefix~\citep{li2021prefix} to the embedded input $x_t$, and the policy $\pi_\theta$ (fine-tuned via LoRA) generates the action $a_t$.

\subsection{Grounded Belief Training}
\label{sec:training}

End-to-end action imitation risks representation collapse, reducing hypotheses to ungrounded task heuristics. To enforce faithful state tracking, \yuhan{our method} decouples physical world modeling from policy optimization, minimizing a joint objective:
\begin{equation}
\mathcal{L} = \mathcal{L}_{\text{act}} + \lambda_{\text{sem}}(\mathcal{L}_{\text{state}} + \mathcal{L}_{\text{out}}) + \lambda_{\text{com}}\mathcal{L}_{\text{com}},
\label{eq:total_loss}
\end{equation}
where $\mathcal{L}_{\text{act}}$ is the standard imitation loss on expert actions $a_t^*$. 

The auxiliary terms ground the belief in physical dynamics: $\mathcal{L}_{\text{state}}$ evaluates Eq.~\eqref{eq:mixture} against ground-truth latent facts to ensure hypotheses span valid world configurations; $\mathcal{L}_{\text{out}}$ trains the belief to anticipate impending transition feedback $p(r_{t+1} \mid B_t, \tilde{a}_t)$, directly supervising the evidence scoring head in Eq.~\eqref{eq:evidence}; and $\mathcal{L}_{\text{com}}$ uses contrastive learning over factually corrupted negatives to enforce semantic invariance in $c_t$.

Finally, stop-gradients on the evidence signal and readout log-weights $\log w_t$ insulate the belief weights from downstream policy gradients, preserving $w_t$ strictly as an environmental likelihood. At inference time, all privileged state supervision is discarded, and \method{} operates solely on online observations and transition feedback.

\section{Experiments}\label{experiments}

\subsection{Experimental Setup}\label{experimental-setup}

\paragraph{Environments.}

\yuhan{We evaluate \method{} on four interactive benchmarks, reporting
SR@30 on ALFWorld~\citep{xi2025agentgym},
task success rate on ScienceWorld~\citep{wang2022scienceworld},
AgentBoard subgoal progress on
BabyAI~\citep{ma2024agentboard},
and BALROG achievement progress on
Crafter~\citep{paglieri2025balrog}.
Environment interfaces and evaluation protocols are detailed
in Appendix~\ref{app:benchmarks}.}

\paragraph{Baselines.}
We group baselines into three categories:
(1)~\emph{\textbf{Open-weight API models }}: Qwen3-8B, Qwen3-32B~\citep{yang2025qwen3}, and GPT-OSS-120B~\citep{openai2025gptoss}, using full interaction histories without task-specific training;
(2)~\emph{\textbf{Proprietary API models}}: GPT-5.6-luna~\citep{openai2026gpt56}, Claude-Haiku-5.5~\citep{anthropic2026haiku55}, and Gemini-3-Flash~\citep{google2025gemini3flash}, also using full interaction histories;
(3)~\emph{\textbf{Benchmark-tuned models}}: fine-tuned Qwen3-8B-FH, BeliefMem~\citep{beliefmem}, AgentEvol~\citep{xi2025agentgym}, PABU~\citep{pabu}, and Agent-BRACE~\citep{agentbrace}.

\paragraph{Training and inference.}

\yuhan{We jointly train the belief modules and LoRA adapters with a
frozen Qwen3-8B backbone; inference updates only the latent belief.
Training details and hyperparameters are provided in
Appendix~\ref{app:training}.}

\subsection{Main Results}\label{main-results}

\yuhan{\textbf{RQ1: How well does \method{} perform?}}

\noindent Table~\ref{tab:overall-performance-across-interactive-environments} shows that \method{} achieves the strongest performance across all four environments among Benchmark-tuned Models, with the largest gains on complex, long-horizon tasks: +22.0 points over PABU on ScienceWorld and +19.55 over AgentEvol on Crafter. Crucially, our 8B agent decisively outperforms full-history prompting on models up to $15\times$ larger (Qwen3-32B, GPT-OSS-120B) as well as frontier proprietary APIs, demonstrating that expanding raw context cannot compensate for a structured belief. Furthermore, the clear advantage over PABU confirms that \yuhan{our method} retains decision-critical details that single natural-language summaries tend to lose. See Appendix~\ref{app:case-study} for a qualitative case study.

\begin{table}[!t]
\centering
\begingroup
\renewcommand{\arraystretch}{0.9}
\setlength{\tabcolsep}{3pt}
\setlength{\aboverulesep}{1pt}
\begin{tabularx}{\textwidth}{X*{4}{rl}}
\toprule
\textbf{Model} & \multicolumn{2}{c}{\hspace*{-1.5em}\shortstack{\textbf{ALFWorld}\\\textbf{SR$\uparrow$}}\hspace*{1.5em}} & \multicolumn{2}{c}{\hspace*{-1.5em}\shortstack{\textbf{ScienceWorld}\\\textbf{SR $\uparrow$}}\hspace*{1.5em}} & \multicolumn{2}{c}{\hspace*{-1.5em}\shortstack{\textbf{BabyAI}\\\textbf{Progress $\uparrow$}}\hspace*{1.5em}} & \multicolumn{2}{c}{\hspace*{-1.5em}\shortstack{\textbf{Crafter}\\\textbf{Progress $\uparrow$}}\hspace*{1.5em}} \\
\midrule
\rowcolor{gray!15}
\multicolumn{9}{c}{\textbf{\textit{Open-weight API Models zero-shot}}} \\
Qwen3-8B-FH {\footnotesize\citep{yang2025qwen3}} & 24.0 & &  1.5 & & 62.16 & &  0.00 & \\
Qwen3-32B-FH {\footnotesize\citep{yang2025qwen3}} & 57.5 & & 17.0 & & 66.73 & & 12.73 & \\
GPT-OSS-120B-FH {\footnotesize\citep{openai2025gptoss}} & 39.5 & & 16.5 & & 64.33 & & 18.64 & \\
\midrule
\rowcolor{gray!15}
\multicolumn{9}{c}{\textbf{\textit{Proprietary API Models}}} \\
GPT-5.6-luna{\footnotesize\citep{openai2026gpt56}}     & 65.50 & & 60.00 & & 73.39 & & 20.00 & \\
Claude-Haiku-5.5{\footnotesize\citep{anthropic2026haiku55}} & 78.50 & & 57.00 & & 81.37 & & 46.82 & \\
Gemini-3-Flash {\footnotesize\citep{google2025gemini3flash}}   & 97.50 & & 72.50 & & 86.32 & & 50.91 & \\
\midrule
\rowcolor{gray!15}
\multicolumn{9}{c}{\textbf{\textit{Benchmark-tuned Models }}} \\
Qwen3-8B-FH {\footnotesize\citep{yang2025qwen3}} &62.50 & & 40.50& & 81.79 & & 6.82 
& \\
BeliefMem {\footnotesize\citep{beliefmem}} &  6.0 & &  1.5 & & 57.46 & &  7.27 & \\
AgentEvol {\footnotesize\citep{xi2025agentgym}} & 88.0 & & 39.5 & & 89.20 & & 39.09 & \\
PABU {\footnotesize\citep{pabu}} & 86.5 & & 61.5 & & 89.57 & & 32.27 & \\
Agent-BRACE {\footnotesize\citep{agentbrace}} & 30.5 & &  28 & & 71.73 & &  6.98 & \\
\rowcolor{blue!10}
\textbf{\method{} (Ours)} & \textbf{92.5} & {\normalfont\footnotesize\textcolor{red}{($\uparrow$4.5)}} & \textbf{83.5} & {\normalfont\footnotesize\textcolor{red}{($\uparrow$22.0)}} & \textbf{89.60} & {\normalfont\footnotesize\textcolor{red}{($\uparrow$0.03)}} & \textbf{58.64} & {\normalfont\footnotesize\textcolor{red}{($\uparrow$19.55)}} \\
\bottomrule
    \end{tabularx}
\endgroup
\caption{Overall performance across interactive environments. ALFWorld reports SR@30 (success within 30 actions). The three blocks contain open-model full-history (FH) baselines, proprietary API full-history baselines, and fine-tuned baselines. Best results within the benchmark-tuned block are in \textbf{bold}. \textcolor{red}{Red} annotations report percentage-point gains over the strongest benchmark-tuned baseline in each \yuhan{interactive environment}.}
\label{tab:overall-performance-across-interactive-environments}
\end{table}

\noindent \yuhan{\textbf{RQ2: Does \method{} reduce inference cost?}}

\noindent \yuhan{Table~\ref{tab:full-history-agent-comparison} reports mean inference
cost over all 200 ALFWorld tasks under a 50-step limit, including
both successful and failed episodes.
Relative to zero-shot Qwen3-8B-FH, \method{} reduces cumulative
input tokens per task by 98.65\% and mean episode length by 66.27\%.
Compared with PABU, it reduces output tokens per task by 57.72\%,
while measured task time is 55.90\% lower.}
This efficiency directly reflects our architectural design: \yuhan{previous methods} must autoregressively decode an explicit textual belief at every decision step, whereas \method{} updates its hypotheses entirely in latent space, decoding only the final action.

\begin{table}[!t]
\centering
\begingroup
\setlength{\belowrulesep}{0.5pt}
\renewcommand{\arraystretch}{0.8}
\setlength{\aboverulesep}{1pt}
\setlength{\tabcolsep}{3pt}
\begin{tabularx}{\textwidth}{l*{4}{>{\centering\arraybackslash}X}}
\toprule
\noalign{\vskip 1pt}
\textbf{Model}
& \shortstack{\textbf{Avg.}\\[-2pt]\textbf{Steps $\downarrow$}}
& \shortstack{\textbf{Input}\\[-2pt]\textbf{Tokens $\downarrow$}}
& \shortstack{\textbf{Output}\\[-2pt]\textbf{Tokens $\downarrow$}}
& \shortstack{\textbf{Task Time}\\[-2pt]\textbf{(s/Task) $\downarrow$}} \\
\midrule
\rowcolor{gray!15}
\multicolumn{5}{c}{\textbf{\textit{API-based Models}}} \\
Llama-3.2-3B-FH  {\footnotesize\citep{meta2024llama32}}     & 48.03 & 332,553 &    741 &  49.10 \\
Qwen3-8B-FH{\footnotesize\citep{yang2025qwen3}}         & 41.24 & 297,154 &    657 &  60.20 \\
Qwen3-32B-FH {\footnotesize\citep{yang2025qwen3}}         & 31.12 & 202,094 &    479 &  27.01 \\
GPT-OSS-120B-FH  {\footnotesize\citep{openai2025gptoss}}     & 37.25 & 244,216 &  3,927 & 198.52 \\
DeepSeek-V4-Flash-FH  {\footnotesize\citep{deepseekai2026v4}} & 18.64 &  81,475 & 13,439 & 132.64 \\
\midrule
\rowcolor{gray!15}
\multicolumn{5}{c}{\textbf{\textit{Benchmark-tuned Models}}} \\
BeliefMem  {\footnotesize\citep{beliefmem}}    & 47.78 & 81,854 & 829 & 23.36 \\
AgentEvol  {\footnotesize\citep{xi2025agentgym}} & 15.15 & 19,185 & 351 & \textbf{12.34} \\
PABU  {\footnotesize\citep{pabu}}& \textbf{13.21} & 5,037 & 499 & 42.02 \\
\rowcolor{blue!10}
\textbf{\method{} (Ours)} & 13.91 & \textbf{4015} & \textbf{211} & 18.53 \\
\bottomrule
\end{tabularx}
\endgroup
\caption{Inference cost on ALFWorld-200 with a 50-step limit. FH denotes full history. The two blocks contain API-based models and benchmark-tuned models; the shaded row highlights \method{}. Steps, token counts, and task times are averaged over all tasks, including failures. Best results are in bold. }
\label{tab:full-history-agent-comparison}
\end{table}

\noindent \yuhan{\textbf{RQ3: How does policy context scale over time?}}

\begin{figure}[!htb]
\centering
\includegraphics[width=0.72\linewidth]{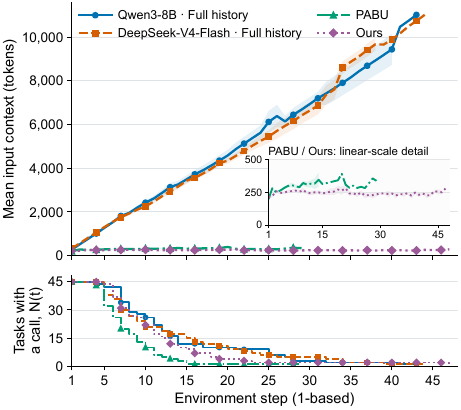}
\caption{Input context length on the 45 ALFWorld tasks successfully completed by four methods under a 50-step budget. Shaded bands denote $\pm 1$ SEM across active tasks at each step; the lower panel reports active task counts. The inset provides detailed scaling for PABU and \method{}.}
\label{fig:context-usage}
\end{figure}

\noindent Total token volume conflates episode length with per-step input size. To isolate how memory representations scale over time, Figure~\ref{fig:context-usage} tracks the policy context length across successive decision steps on the 45 ALFWorld tasks successfully solved \yuhan{by all four displayed methods.} Input context encompasses all tokens fed to the policy before generating an action, including task instructions, current observations, admissible actions, and history or belief representations.

\yuhan{As interactions accumulate, full-history models require increasingly
long policy inputs, whereas \method{} maintains an approximately
stable context.
At decision step 10, \method{} reduces mean policy input length by
approximately 89\% relative to Qwen3-8B-FH and 88\% relative to
DeepSeek-V4-Flash-FH.
By step 20, the reduction reaches approximately 94\% for both
baselines, illustrating the growing savings over longer interactions.}

These dynamics demonstrate that recursive latent belief tracking eliminates cumulative context overhead while preserving decision-critical state over extended episodes.

\section{Ablation Studies and Analysis}\label{ablation-studies-and-analysis}

\begin{table}[!htb]
\centering
\small
\setlength{\tabcolsep}{2pt}
\begin{tabularx}{0.8\linewidth}{>{\raggedright\arraybackslash}Xrrrr}
\toprule
\textbf{Variant} & \shortstack[r]{\textbf{SR@30}\\\textbf{(\%) $\uparrow$}} & \shortstack[r]{\textbf{SR@50}\\\textbf{(\%) $\uparrow$}} & \textbf{$\Delta_{50}$} & \shortstack[r]{\textbf{Avg.}\\\textbf{Steps $\downarrow$}} \\
\midrule
\rowcolor{ablationhl}
\textbf{Full Model}         & \textbf{92.5} & \textbf{95.5} &   0.0 & \textbf{13.91} \\
Unstructured Latent  & 46.5 & 51.5 & $-44.0$ & 31.40 \\
w/o Delta                   & 86.0 & 90.5 & $-5.0$ & 17.04 \\
w/o Common                  & 91.5 & 93.5 & $-2.0$ & 15.38 \\
w/o Belief Input     & 80.0 & 82.0 & $-13.5$ & 19.13 \\
w/o $\mathcal{L}_{\text{state}}$      & 86.5 & 92.0 & $-3.5$ & 16.71 \\
w/o $\mathcal{L}_{\text{out}}$   & 88.5 & 94.0 & $-1.5$ & 16.17 \\
\bottomrule
\end{tabularx}
\caption{Component ablations on 200 ALFWorld tasks under identical training and evaluation protocols. SR@30 and SR@50 measure success within 30 and 50 actions, respectively. $\Delta_{50}$ denotes the percentage-point change in SR@50 relative to the full model; average steps include successful and failed tasks under the 50-step limit. }
\label{tab:component-ablations-on-alfworld}
\end{table}

\subsection{Component Ablations}\label{component-ablation-studies}

\begin{figure}[!t]
\centering
\includegraphics[width=0.7\linewidth]{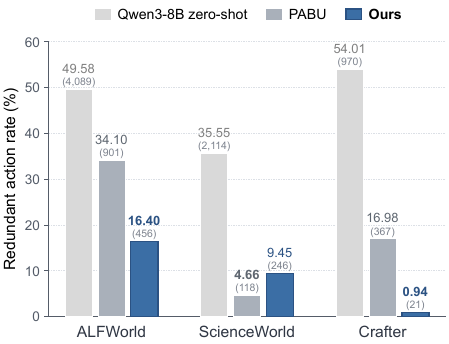}
\caption{Redundant action rates across three environments (\%, lower is better), with absolute counts in parentheses, evaluated over all executed actions across successful and failed tasks.}
\label{fig:redundant-actions-baselines}
\end{figure}

Table~\ref{tab:component-ablations-on-alfworld} compares six retrained variants. Replacing the structured Common--Delta latent (4 Common + 8 Delta tokens) with 12 unstructured latent tokens, while retaining the same policy prefix length and auxiliary losses, reduces SR@30 from 92.5\% to 46.5\% and SR@50 from 95.5\% to 51.5\%, with average steps increasing to 31.40. This gap supports the effectiveness of the proposed representation and update structure over the evaluated unstructured alternative. Removing belief tokens from the policy input (\textbf{w/o  Belief Input}) reduces SR@50 to 82.0\%. Within the architecture, removing Delta or Common reduces SR@50 by 5.0 or 2.0 points, respectively, supporting their complementary roles.
\yuhan{What's more,} auxiliary supervision also contributes: removing state supervision reduces SR@50 to 92.0\% and increases average steps from 13.9 to 16.7. Removing observation prediction reduces SR@30 from 92.5\% to 88.5\%, indicating its contribution to early task completion.
\subsection{Redundant Action Analysis}\label{redundant-action-analysis}

Evidence degradation manifests in behavioral artifacts before it lowers task success: an agent re-inspects visited locations, searches for objects it has already found, or retries actions that recently failed. We utilize redundant actions as an empirical probe to evaluate what the belief retains, defining redundancy by contextual task relevance instead of literal command repetition. Specifically, an action is deemed redundant if, given past observations and environment rules, it contributes neither to advancing pending subgoals nor to resolving necessary task uncertainties.
We evaluate Qwen3-8B zero-shot, PABU, and \method{} across ALFWorld, ScienceWorld, and Crafter (Figure~\ref{fig:redundant-actions-baselines}). On ALFWorld and Crafter, \method{} achieves the lowest redundancy rates across all methods (16.40\% and 0.94\%), outperforming PABU by 17.70 and 16.04 percentage points, and improving over zero-shot prompting by more than 33 points. 
\yuhan{In addition,} on ScienceWorld, \method{} cuts zero-shot redundancy from 35.55\% to 9.45\%. While PABU records a lower numerical redundancy rate (4.66\%), this metric must be contextualized with task completion: PABU succeeds on only 61.5\% of ScienceWorld tasks, terminating prematurely on complex trajectories where redundant behaviors typically accumulate. In contrast, \method{} achieves 83.5\% task success (+22.0 points over PABU) while maintaining single-digit redundancy. These results demonstrate that structured latent particles systematically curtail aimless interactions across complex long-horizon environments.

\paragraph{Fine-grained error breakdown.}
To identify which memory failures our latent belief resolves, we categorize ALFWorld redundant actions into five mutually exclusive failure modes: \emph{re-observation}, \emph{unnecessary search}, \emph{unnecessary manipulation}, \emph{failed retries}, and \emph{repeated operations} (formal definitions in Appendix~\ref{app:error_categories}). 
And Figure~\ref{fig:alfworld-redundancy-breakdown} visualizes the distribution of these failure modes across executed actions. While zero-shot Qwen3-8B exhibits high redundancy across all modes, PABU curbs re-observations but suffers severely from failed retries, which account for 20.06\% of its total actions. This reveals an inherent limitation of natural-language summaries: textual memory typically logs that an action was executed, but struggles to retain whether it failed or why.

In contrast, \method{} completely eliminates failed retries, unnecessary manipulations, and repeated operations (0\% across all three categories). Although our re-observation rate is slightly higher, eradicating the 20-point retry failure mode reduces overall redundancy by 17.70 points. This decisive gain confirms that our observation-conditioned particle update effectively propagates failure feedback through negative evidence signals. Meanwhile, re-inspecting known information remains the primary residual source of redundancy.

\begin{figure}[!t]
\centering
\includegraphics[width=0.6\linewidth]{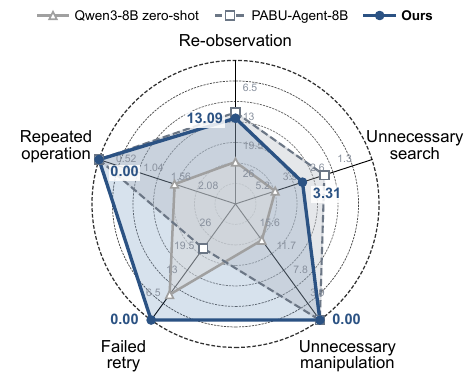}
\caption{ALFWorld redundant-action breakdown across five mutually exclusive categories (percentage of all executed actions, lower is better). Blue annotations mark \method{}'s rates.}
\label{fig:alfworld-redundancy-breakdown}
\end{figure}

\subsection{Observation Retention Analysis}\label{sec:observation-retention}

A textual belief selects observation details for its summary, whereas Common maintains a fixed-capacity latent state through recursive updates. We evaluate whether previously observed information remains recoverable after these updates through a diagnostic observation-recall task.

To isolate memory retention from policy divergence, all methods replay identical ALFWorld action--observation sequences where an agent visits a location and subsequently executes up to eight actions that never mention that location again. At lags of 0, 1, 2, 4, and 8 steps, we train a separate linear probe for each method and lag to identify the original observation among four candidates (25\% chance baseline). Lag 0 is measured immediately after the target observation has been incorporated into memory. For this vector-based diagnostic, textual baselines (AgentEvol's raw history and PABU's summary) are encoded using frozen Qwen3-8B embeddings and mean-pooled into vectors of identical dimensionality before probing, while our method is probed directly from the Common state vectors.

\begin{table}[!ht]
\centering
{\small%
\begin{tabular}{lrrrrr}
\toprule
\textbf{Method} & \textbf{0 steps} & \textbf{1 step} & \textbf{2 steps} & \textbf{4 steps} & \textbf{8 steps} \\
\midrule
AgentEvol & 78.13 & 66.67 & 62.50 & 52.60 & 42.19 \\
PABU      & 25.00 & 25.00 & 25.52 & 25.00 & 25.00 \\
\rowcolor{retentionhl}
\textbf{Ours} & \textbf{99.48} & \textbf{99.48} & \textbf{98.96} & \textbf{98.96} & \textbf{98.44} \\
\bottomrule
\end{tabular}%
}
\caption{Observation identification (\%, higher is better) across subsequent interaction lags on 48 ALFWorld trajectories, evaluated using vector-based 4-way classification probes. }
\label{tab:common-evidence-recall}
\end{table}

Table~\ref{tab:common-evidence-recall} shows that Common maintains high observation-identification accuracy, decreasing only from 99.48\% to 98.44\% across the tested lags. PABU remains near chance, while AgentEvol declines from 78.13\% to 42.19\%. These results measure recovery through the vector probes; we further examine the textual memories using a direct text reader in Appendix~\ref{app:candidate-assisted-recovery}.
And appendices~\ref{delta-token-overlap} and~\ref{particle-count-token-budget} further analyze the diversity of Delta particles and the effect of particle count and token budget.

\section{Conclusion}\label{conclusion}

\yuhan{We introduced \method{}, a recurrent latent belief framework that
factorizes memory into a shared state and weighted hypotheses,
updated using action-outcome evidence.
With a fixed token belief prefix, \method{} achieves competitive
performance across four interactive benchmarks and reduces token
consumption on ALFWorld.
These results support structured latent beliefs as a compact and
effective memory representation for long-horizon language agents. 

We hope this work encourages future memory architectures trained to actively predict the environment instead of passively transcribing the past.}

\clearpage

\bibliographystyle{plainnat}
\bibliography{particlebelief}

\appendix

\section{Effect of Delta Particles}\label{delta-token-overlap}

To verify that multi-hypothesis particles avoid representational collapse, this appendix investigates whether individual Delta tokens yield distinct vocabulary-space projections when processed by the frozen language model in isolation.

\paragraph{Setup.}
For each evaluated state, we feed each delta token to the backbone individually at the same input position. We omit task instructions, the Common prefix, and other Delta tokens, ensuring that output differences stem only from the particle embeddings themselves. 

Following the logit lens paradigm~\citep{nostalgebraist2020logitlens}, we decode hidden states at every layer through the frozen final normalization and unembedding matrix, recording the ten most probable vocabulary tokens. Letting $S_i^{(l)}$ denote this top-10 set for particle $i$ at layer $l$, we compute the mean pairwise overlap across all $\binom{K}{2}$ particle pairs:
\begin{equation}
O^{(l)} = \frac{1}{\binom{K}{2}}\sum_{i<j}\frac{|S_i^{(l)}\cap S_j^{(l)}|}{10}.
\label{eq:delta-top10-overlap}
\end{equation}
Here, $O^{(l)}=1$ indicates identical top-10 vocabulary selections across particles, while $O^{(l)}=0$ denotes completely disjoint sets. Overlap values are averaged over all $\binom{8}{2}=28$ pairs within each state, across 12 representative diagnostic states.

\begin{figure}[t]
\centering
\includegraphics[width=0.65\linewidth]{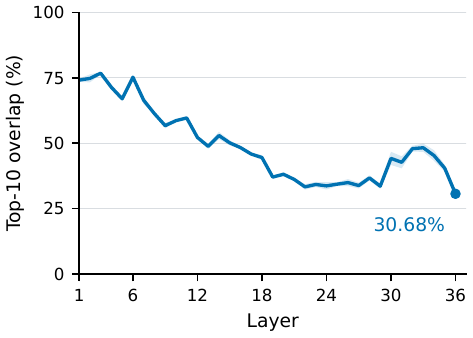}
\caption{Layer-wise top-10 token overlap across Delta particles, averaged over 28 particle pairs across 12 diagnostic states. Final-layer overlap drops to 30.68\%.}
\label{fig:delta-token-overlap}
\end{figure}

\paragraph{Results.}
Figure~\ref{fig:delta-token-overlap} illustrates that token overlap declines from 74.17\% at layer 1 to 30.68\% at the final layer. While intermediate layers exhibit non-monotonic fluctuations, particles progressively diverge across the transformer depth. At the output layer, two particles share only about three of their ten most probable tokens on average, indicating that the Delta particles retain distinct vocabulary-space projections and successfully mitigate representational collapse.

\section{Particle Count and Token Budget}\label{particle-count-token-budget}

We investigate how agent performance scales when the number of Common readout tokens and Delta particles are varied jointly.  We compare three configurations: $2{+}4$, $4{+}8$, and $8{+}16$, corresponding to total prefix budgets of 6, 12, and 24 tokens, with $4{+}8$ serving as the default setting. Each variant is trained on the raw 7,200/800 train/dev source split without multi-view trajectory augmentations. Models are trained independently from the Qwen3-8B backbone and evaluated on the same 200 ALFWorld test tasks. 

\begin{table}[t]
\centering
{\small%
\begin{tabular}{rrrrrr}
\toprule
\shortstack[r]{\textbf{Common}\\\textbf{Tokens}} & \shortstack[r]{\textbf{Delta}\\\textbf{Particles}} & \shortstack[r]{\textbf{Total}\\\textbf{Prefix}} & \shortstack[r]{\textbf{SR@30}\\\textbf{(\%) $\uparrow$}} & \shortstack[r]{\textbf{SR@50}\\\textbf{(\%) $\uparrow$}} & \shortstack[r]{\textbf{Avg.}\\\textbf{Steps $\downarrow$}} \\
\midrule
2 & 4  & 6  & 87.5 & 92.0 & 16.39 \\
\rowcolor{ablationhl}
4 & 8  & 12 & \textbf{89.0} & \textbf{94.0} & \textbf{15.78} \\
8 & 16 & 24 & 88.0 & 93.5 & 16.09 \\
\bottomrule
\end{tabular}%
}
\caption{ALFWorld performance under varying Common and Delta token budgets. All variants are trained on the unaugmented 7,200/800 source split. The green-shaded row denotes our default configuration. Bold marks the best result per column.}
\label{tab:particle-count-token-budget}
\end{table}

Table~\ref{tab:particle-count-token-budget} indicates that policy performance saturates quickly with respect to the belief budget. Among the three evaluated configurations, the default $4{+}8$ budget achieves the highest SR@30 (89.0\%) and SR@50 (94.0\%) while requiring the fewest average steps (15.78). Constraining capacity to $2{+}4$ lowers SR@50 by 2.0 points, demonstrating that an undersized particle set lacks sufficient expressive capacity to maintain competing hypotheses. Conversely, expanding the budget to $8{+}16$ yields no performance benefit (93.5\% SR@50) despite doubling the prefix footprint.

\section{Redundant Action Categories}\label{app:error_categories}

An action is \emph{redundant} if it (i) reveals no new information, (ii) causes no task-relevant state change, and (iii) satisfies no precondition for a later step.  Redundancy is judged only from information available before execution: exploring unverified states is not considered redundant merely because a more direct trajectory existed in retrospect. Under this definition, redundancy encompasses both literal string repetitions and non-repetitive detours made avoidable by available context. 

We classify redundant actions into five mutually exclusive categories.  \emph{Re-observation} gathers facts the agent has already observed. \emph{Unnecessary search} continues exploring after the target location is established. \emph{Unnecessary manipulation} performs state-changing actions that are irrelevant or destructive to the task. \emph{Failed retry} re-attempts an unachieved objective without addressing the cause of its failure. \emph{Repeated operation} re-applies an effect that has already been achieved and still holds.Figure~\ref{fig:alfworld-redundancy-breakdown} visualizes their rates on ALFWorld.

\paragraph{Category boundaries.}
Three rules resolve actions that could fit two categories:
\textbf{(1)}~\emph{Re-observation vs.\ unnecessary search} depends on whether the target location is known. Searching a previously eliminated container while the target is still unknown is a re-observation; searching other containers after the target has been found is an unnecessary search.
\textbf{(2)}~\emph{Failed retry vs.\ repeated operation} depends on whether the intended effect has already been achieved. Re-opening a locked door without acquiring the key is a failed retry; opening a door that is already open is a repeated operation.
\textbf{(3)}~\emph{Unnecessary manipulation vs.\ repeated operation} depends on whether the action disrupts progress. Removing an item already placed in its correct receptacle is an unnecessary manipulation; reheating an object that is already heated is a repeated operation.

\paragraph{Non-redundant repetition.}
Repetition at the string level does not inherently indicate redundancy. Multi-object tasks legitimately require consecutive \emph{pick} actions; combat, mining, or waiting sequences frequently demand successive identical operations; and re-inspecting an environment following a state-altering action is often necessary. Furthermore, an initial failure followed by a retry after resolving the prerequisite condition is strictly classified as valid task execution.

\section{Hyperparameter Details}\label{app:training}

Table~\ref{tab:hyperparameters} summarizes the hyperparameter configuration for the main experiments.

\begin{table}[t]
\centering
\small
\renewcommand{\arraystretch}{0.95}
\setlength{\tabcolsep}{4pt}
\begin{tabularx}{0.7\linewidth}{@{}>{\raggedright\arraybackslash}X>{\raggedright\arraybackslash}p{2.85cm}@{}}
\toprule
\textbf{Hyperparameter} & \textbf{Value} \\
\midrule
\multicolumn{2}{@{}l}{\textit{Backbone \& Adaptation}} \\
Policy model & Qwen3-8B (frozen) \\
LoRA rank $r$ / $\alpha$ / dropout & 32 / 64 / 0.05 \\
Max sequence length & 2,048 (+12 prefix) \\
Backbone precision & BF16 \\
\midrule
\multicolumn{2}{@{}l}{\textit{Belief Representation}} \\
Common / Delta state dim $d$ & 4,096 \\
Delta particle count $K$ & 8 \\
Common / Delta prefix tokens & 4 / 8 (12 total) \\
Initialization noise scale $\sigma$ & 0.01 \\
Particle adapter bottleneck & 64 (scale 1.0) \\
Mixture-gate hidden dim & 512 \\
Belief module precision & FP32 \\
\midrule
\multicolumn{2}{@{}l}{\textit{Optimization}} \\
Optimizer & AdamW \\
AdamW $\beta_1$ / $\beta_2$ & 0.9 / 0.999 \\
AdamW $\beta_1$ / $\beta_2$ & 0.9 / 0.999 \\
AdamW $\epsilon$ / weight decay & $10^{-8}$ / 0.01 \\
Learning rate / schedule & $2\times10^{-5}$ / linear \\
Warmup ratio / epochs & 0.03 / 3 \\
Effective batch size & 16 episodes \\
Gradient clipping norm & 1.0 \\
BPTT truncation window & 1 env.\ step \\
\midrule
\multicolumn{2}{@{}l}{\textit{Auxiliary Objectives}} \\
State loss weight $\lambda_{\mathrm{sem}}$ & 0.2 \\
Contrastive loss weight $\lambda_{\mathrm{com}}$ & 0.1 \\
Max state queries per step & 6 \\
Contrastive samples & 1 pos.\ / 31 neg. \\
Contrastive temperature & 1.0 \\
\bottomrule
\end{tabularx}
\caption{Hyperparameter settings for the main experiments.}
\label{tab:hyperparameters}
\end{table}

\section{Joint Prediction of Latent Facts}\label{app:joint-prediction}

\begin{figure}[t]
\centering
\includegraphics[width=0.75\linewidth]{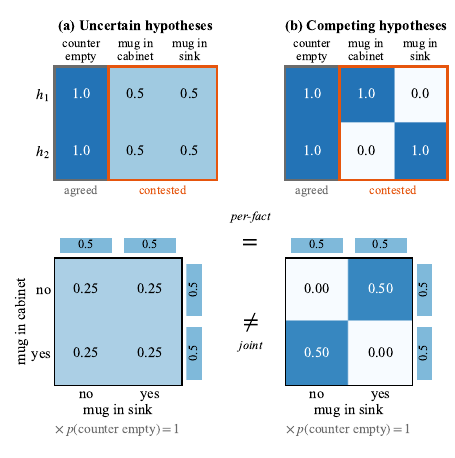}
\caption{\textbf{Joint vs.\ per-fact prediction (Eq.~\ref{eq:mixture}).} With countertop~1 known to be empty, the mug is in either cabinet~1 or sink~1. \textbf{Top:} per-hypothesis predictions $p_\phi(y_q \mid c_t, \delta_t^i)$ for (a) uncertain and (b) competing hypotheses. \textbf{Bottom:} the resulting joint distributions. Both cases share the same per-fact marginals (0.5), but only (b) assigns zero probability to the mug being in both or neither location. The agreed fact (gray) factors out as a shared multiplier.}
\label{fig:joint}
\end{figure}

In Section~\ref{sec:formulation}, we formulate the prediction of $Q$ latent facts as a mixture of factorized distributions (Eq.~\ref{eq:mixture}):
\begin{equation}
p(y_{1:Q} \mid B_t) = \sum_{i=1}^K w_t^i \prod_{q=1}^Q p_\phi(y_q \mid c_t, \delta_t^i).
\label{eq:app_joint}
\end{equation}
We further explain why placing the product inside the mixture couples the latent facts and how this motivates the Common--Delta split.

\subsection{Coupled Joint Prediction vs.\ Independent Factorization}
Predicting each fact independently instead moves the product outside the mixture:
\begin{equation}
p_{\text{indep}}(y_{1:Q} \mid B_t) = \prod_{q=1}^Q \left[ \sum_{i=1}^K w_t^i p_\phi(y_q \mid c_t, \delta_t^i) \right].
\label{eq:app_indep}
\end{equation}
A marginal of $0.5$ can arise from hypotheses that are each uncertain, or from hypotheses that are each certain but disagree (Figure~\ref{fig:joint}(a,b)). Eq.~\eqref{eq:app_indep} assigns both cases the same joint distribution and thus ignores mutual exclusivity between facts.

\subsection{Numerical Walkthrough of the Toy Example}
Consider a mug that is in either the cabinet ($y_{\text{cab}}$) or the sink ($y_{\text{sink}}$), with two equally weighted hypotheses ($w^1 = w^2 = 0.5$). We write joint states as pairs $(y_{\text{cab}}, y_{\text{sink}})$.

\paragraph{Case (a): Uncertain hypotheses.}
Both hypotheses predict $p(y_{\text{cab}}{=}1) = p(y_{\text{sink}}{=}1) = 0.5$, so Eq.~\eqref{eq:app_joint} gives
\begin{equation}
p(1, 1) = p(0, 0) = 0.5(0.5^2) + 0.5(0.5^2) = 0.25,
\end{equation}
assigning 25\% to the mug being in both places and 25\% to it being in neither.

\paragraph{Case (b): Competing hypotheses (\method{}).}
Each particle commits to one configuration: $h_1$ places the mug in the cabinet $(1, 0)$ and $h_2$ in the sink $(0, 1)$. Eq.~\eqref{eq:app_joint} then gives
\begin{align}
p(1, 1) = p(0, 0) &= 0.5(1{\cdot}0) + 0.5(0{\cdot}1) = \mathbf{0.00}, \label{eq:impossible_states} \\
p(1, 0) = p(0, 1) &= 0.5(1{\cdot}1) + 0.5(0{\cdot}0) = \mathbf{0.50}, \label{eq:valid_states}
\end{align}
ruling out both impossible states. Eq.~\eqref{eq:app_indep} would instead average the marginals to $0.5$ and again assign $0.25$ to each impossible state.

\subsection{Algebraic Separation of Agreed and Contested Facts}\label{app:algebraic-separation}
Suppose the facts split into agreed facts $\mathcal{A}$, on which all hypotheses agree, $p_\phi(y_a \mid c_t, \delta_t^i) = p_\phi(y_a \mid c_t)$, and contested facts $\mathcal{C}$. The agreed terms then factor out of the mixture:
\begin{equation}
\begin{aligned}
p(y_{1:Q} \mid B_t) &= \left[ \prod_{a \in \mathcal{A}} p_\phi(y_a \mid c_t) \right] \\
&\quad \times \sum_{i=1}^K w_t^i \prod_{c \in \mathcal{C}} p_\phi(y_c \mid c_t, \delta_t^i).
\end{aligned}
\label{eq:factor_out}
\end{equation}
The consensus becomes a shared multiplier (gray in Figure~\ref{fig:joint}), and uncertainty is confined to the mixture over contested facts (orange). This motivates storing confirmed evidence in the common state $c_t$ and reserving the deltas $\delta_t^i$ for alternatives that still diverge.

\section{Benchmark Details}\label{app:benchmarks}

\textbf{ALFWorld}~\citep{xi2025agentgym} is a text-based household environment in which agents locate and manipulate objects across rooms, often through multi-stage operations such as cleaning, heating, and cooling. The AgentGym evaluation setting includes 200 test tasks with a 50-action limit. Performance is measured by SR@30 and SR@50, the success rates within the first 30 and 50 actions, respectively.

\noindent\textbf{BabyAI}~\citep{ma2024agentboard} tests grounded instruction following in partially observable grid worlds, requiring agents to track spatial layout, inventory, and prerequisites such as finding a key before opening a locked door. The AgentBoard evaluation setting includes 112 test instances across 28 levels, with a 30-action limit. Performance is measured by mean subgoal progress.

\noindent\textbf{Crafter}~\citep{paglieri2025balrog} is a procedurally generated open-world survival game with 22 achievements covering resource gathering, crafting, and combat, requiring agents to track vital stats, inventory, and prerequisite chains. Evaluation uses the BALROG text interface and evaluator on ten held-out seeds, with up to 2,000 steps per episode. Performance is measured by achievement progress, the fraction of achievements unlocked averaged over seeds.

\noindent\textbf{ScienceWorld}~\citep{wang2022scienceworld} is a text environment for multi-step scientific experiments, such as measuring melting points, building circuits, and classifying organisms. Because standard subsets are dominated by short-horizon tasks, we evaluate on 200 longer-horizon variations from 19 task categories of the official unseen test split, with a 30-decision limit. We use the official easy setting, which removes navigation and door- or container-opening steps but leaves the experimental procedure unchanged, so agents must still retain observations across multiple steps to complete each task. A task succeeds only if it reaches the maximum score of 100.

\section{Memory Recovery with Direct Text Reading}
\label{app:candidate-assisted-recovery}

While vector probes may not fully reflect the recoverability of textual memories, we use a frozen Qwen3-8B text reader for Full history and PABU. All methods receive the same four complete observation candidates on 24 paired branches that differ in target-object presence and share identical subsequent updates. Observation identification and target-object presence derived from the selected observation are scored separately.

\begin{table}[!ht]
\centering
{\small%
\begin{tabular}{lrrr}
\toprule
\textbf{Method} &
\textbf{Obs., lag 0} &
\textbf{Obs., lag 8} &
\textbf{Fact, lag 8} \\
\midrule
Full history & \textbf{100.00} & \textbf{100.00} & \textbf{100.00} \\
PABU         & 41.67  & 33.33  & 50.00 \\
\rowcolor{retentionhl}
\textbf{Ours} & \textbf{100.00} & \textbf{100.00} & \textbf{100.00} \\
\bottomrule
\end{tabular}%
}
\caption{Candidate-assisted recovery (\%) on 24 paired observation branches from 12 base trajectories. Obs.: identification among four complete observations. Fact: target-object presence derived from the selected observation. The shaded row highlights \method{}; best results (including ties) are in bold.}
\label{tab:candidate-assisted-recall}
\end{table}

Table~\ref{tab:candidate-assisted-recall} shows that Full history and Common achieve 100\% observation identification and fact accuracy after eight updates, whereas PABU achieves 33.33\% and 50.00\%, respectively. At lag 8, the text reader selects the same candidate for both members of every PABU observation pair, failing to distinguish whether the target object was previously seen. These results show that target facts remain recoverable from Common after eight recursive updates under candidate-assisted reading, matching Full history and outperforming PABU.

\section{Case Study}
\label{app:case-study}

Figures~\ref{fig:case-study-2547} and~\ref{fig:case-study-2547-cont} present the complete 15-step ALFWorld episode of \method{}, with detailed Common--Delta state descriptions at three key decisions.

\begin{figure}[p]
\small
\setlength{\parindent}{0pt}
\input{figures/case_study_2547}
\caption{In this ALFWorld case-study, the Common component retains the lamp location on dresser~1, while Delta hypotheses shift away from drawers confirmed empty toward uninspected locations. Once the bowl is found on desk~1, the agent returns to the lamp and uses it to examine the bowl, ultimately completing the task successfully. Cross-hatched boxes denote latent belief components.}
\label{fig:case-study-2547}
\end{figure}

\begin{figure}[p]
\small
\setlength{\parindent}{0pt}
\input{figures/case_study_2547_continued}
\caption{Continuation of Figure~\ref{fig:case-study-2547}.}
\label{fig:case-study-2547-cont}
\end{figure}

\end{document}